%% file: main.tex
\documentclass{article}
\usepackage[12pt]{extsizes}
\usepackage[utf8]{inputenc}
\usepackage{amsmath}
\usepackage{amssymb}
\usepackage{amsthm}
\usepackage{natbib}
\usepackage{graphicx}
\usepackage{lipsum}
\usepackage[dvipsnames]{xcolor}
\usepackage[hidelinks]{hyperref}
\hypersetup{
    colorlinks = true,
    linkcolor={blue!50!black},
    citecolor={blue!50!black},
    urlcolor={blue!80!black}
}
\usepackage{authblk}
\usepackage{geometry}
\usepackage{mathtools}
\usepackage{amsfonts}
\usepackage{relsize}
\usepackage{multirow}
\usepackage{booktabs}
\usepackage{tabularx}
\usepackage{listings}

\usepackage{xspace}
\usepackage[ruled,vlined]{algorithm2e}

\usepackage{multirow}

\newcommand{\mname}{\text{SkipGNN}\xspace}

\newtheorem*{problem*}{Problem}

\usepackage[p,osf]{cochineal}
\usepackage[scale=.95,type1]{cabin}
\usepackage[cochineal,bigdelims,cmintegrals,vvarbb]{newtxmath}
\usepackage[zerostyle=c,scaled=.94]{newtxtt}
\usepackage[cal=boondoxo]{mathalfa}
\usepackage{microtype}

\usepackage{etoolbox}
\apptocmd{\thebibliography}{\raggedright}{}{}

\input{math_commands.tex}


\title{DeepPurpose: A Deep Learning Library for Drug-Target Interaction Prediction}

\author[1]{Kexin Huang}
\author[2]{Tianfan Fu}
\author[3]{Lucas M. Glass}
\author[1]{Marinka Zitnik}
\author[3]{Cao Xiao}
\author[4]{Jimeng Sun}

\affil[1]{Harvard University, Boston, MA}
\affil[2]{Georgia Institute of Technology, Atlanta, GA}
\affil[3]{IQVIA, Cambridge, MA}
\affil[4]{University of Illinois at Urbana-Champaign, Urbana, IL}

\begin{document}
\maketitle

\begin{abstract}
    Accurate prediction of drug-target interactions (DTI) is crucial for drug discovery. Recently, deep learning (DL) models for show promising performance for DTI prediction. However, these models can be difficult to use for both computer scientists entering the biomedical field and bioinformaticians with limited DL experience. We present DeepPurpose, a comprehensive and easy-to-use deep learning library for DTI prediction. DeepPurpose supports training of customized DTI prediction models by implementing 15 compound and protein encoders and over 50 neural architectures, along with providing many other useful features. We demonstrate state-of-the-art performance of DeepPurpose on several benchmark datasets. \\
    \textbf{Supplementary}: Supplementary data are available at \href{https://oup.silverchair-cdn.com/oup/backfile/Content_public/Journal/bioinformatics/PAP/10.1093_bioinformatics_btaa1005/1/btaa1005_supplementary_data.zip?Expires=1610334301&Signature=ZhtbbbpoCcq0hMbPH9pgvUFiBEO69A0sx3UED~esplEHPPHhuTc87rfbxeEAR4wPljjAS~bLwaIc0h4gihhs4NzDQ9optPkaNOVTHO-yUrm1wyQgCZuvAy8Ok6zTjWFXO1dCZA2NYsDTbvvE3TPv7EyXUcAUJcSf4S2HihssRYSGh5r8~OORcbva6Ah5NKHG6sSShQoUIBU6TSObcR1owq956PkoR-WFtzZxP7wANfunGtqHQILHatojnS0aM1pKhCryg9jPLG~SKDFCa4Qm9rYoSyo9WezwsvHQTsVsd~UF8h2E~2Oq5QDXIjHIQiMcimJ5k-KGlz~F-JnF-czvDQ__&Key-Pair-Id=APKAIE5G5CRDK6RD3PGA}{Bioinformatics online}.\\

    \textbf{Acknowledgement}: This article has been accepted for publication in Bioinformatics \copyright 2020. Published by Oxford University Press. All rights reserved.
\end{abstract}
\section{Introduction}

Drug-target interactions (DTI) characterize the binding of compounds to protein targets~\citep{santos2017comprehensive}. Accurate identification of molecular drug targets is fundamental for drug discovery and development~\citep{rutkowska_modular_2016,zitnik_machine_2019} and is especially important for finding effective and safe treatments for new pathogens, including SARS-CoV-2~\citep{velavan_covid19_2020}.

Deep learning (DL) has advanced traditional computational modeling of compounds by offering an increased expressive power in identifying, processing, and extrapolating complex patterns in molecular data~\citep{ozturk_deepdta_2018,lee_deepconv-dti_2019}. There are many DL models designed for DTI prediction~\citep{ozturk_deepdta_2018,lee_deepconv-dti_2019,nguyen_graphdta_2020}. However, to generate predictions, deploy DL models in practice, test, and evaluate model performance, one needs considerable programming skills and extensive biochemical knowledge. Prevailing tools are designed for experienced interdisciplinary researchers. They are challenging to use by both computer scientists entering the biomedical field and domain bioinformaticians with limited experience in training and deploying DL models. Furthermore, each open-sourced tool has a different programming interface and is coded differently, which prevents easy integration of outputs from various methods for model ensembles~\citep{yang_analyzing_2019}. 

Here, we introduce DeepPurpose, a deep learning library for encoding and downstream prediction of proteins and compounds. DeepPurpose allows rapid prototyping via a programming framework that implements over 50 deep learning models, seven protein encoders, and eight compound encoders. Empirically, we find that models implemented in DeepPurpose achieve state-of-the-art prediction performance on DTI benchmark datasets.

\section{DeepPurpose Library}

\begin{figure*}[t]
    \centering
    \includegraphics[width = \textwidth]{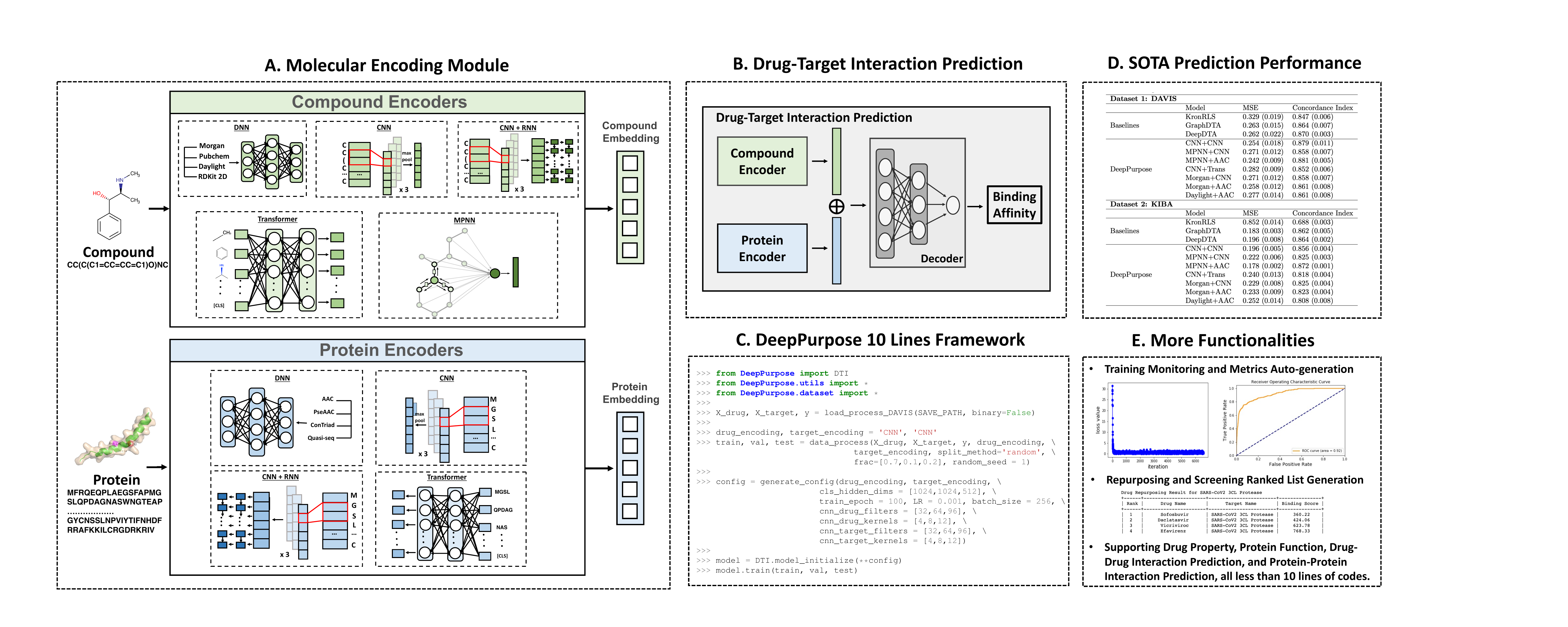}
    \caption{\textbf{Overview of DeepPurpose library.} (A) DeepPurpose takes as input the SMILES of a compound and a protein's amino acid sequence and then generates embeddings for them. (B) The learned embeddings are then concatenated and fed into a decoder to predict DTI binding affinity. (C) DeepPurpose provides a simple but flexible programming framework that implements over 50 state-of-the-art deep learning models for DTI prediction. (D) DeepPurpose models achieve comparable performance with three other DTI prediction algorithms on two benchmark datasets. (E) Finally, DeepPurpose has many functionalities, including monitoring the training process, debugging, and generation ranked lists for repurposing and screening. Further, DeepPurpose supports other downstream prediction tasks (e.g., drug-drug interaction prediction, compound property prediction).}
    \label{fig:my_label}
\end{figure*}

Deep learning models for DTI prediction can be formulated as an encoder-decoder architectures~\citep{cho_properties_2014}. DeepPurpose library implements a unifying encoder-decoder framework, which makes the library uniquely flexible. By merely specifying an encoder's name, the user can automatically connect a encoder of interest with the relevant decoder. DeepPurpose then trains the corresponding encoder-decoder model in an end-to-end manner. Finally, the user accesses the trained model either programmatically or via a visual interface and uses the model for DTI prediction.

\subsection{Module for Encoding Proteins and Compounds} 

DeepPurpose takes the compound's simplified molecular-input line-entry system (SMILES) string and protein amino acid sequence pair as input. Then, they are fed into molecular encoders which specifies a deep transformation function that maps compounds and proteins to a vector representation. In particular, for compounds, DeepPurpose provides eight encoders using different modalities of compounds: Multi-Layer Perceptrons (MLP) on Morgan, PubChem, Daylight and RDKit 2D Fingerprint; Convolutional Neural Network (CNN) on SMILES strings; Recurrent Neural Network (RNN) on top of CNN; transformer encoders on substructure fingerprints; message passing graph neural network on molecular graph. For proteins, DeepPurpose provides seven encoders for the input amino acid sequence: MLP on Amino Acid Composition (AAC), Pseudo AAC, Conjoint Triad, Quasi-Sequence descriptors; CNN on amino acid sequences; RNN on top of CNN; transformer encoder on substructure fingerprints. Note that alternative input features may not work for a specific encoder architecture. The detailed encoder specifications and references are described in Supplementary.

\subsection{Module for DTI Prediction} 

DeepPurpose feeds the learned protein and compound embeddings into an MLP decoder to generate predictions. Output scores include both continuous binding scores, such as the median inhibitory concentration ($\textrm{IC}_{50}$), as well as binary outputs indicating whether a protein binds to a compound. The library detects whether the task is regression or classification and switches to the correct loss function and evaluation metrics. In the case of regression, we use the Mean Square Error (MSE) as the loss function and MSE, Concordance Index, and Pearson Correlation as performance metrics. In the classification case, we use Binary Cross Entropy as the loss function and Area Under the Receiver Operating Characteristics (AUROC), Area Under Precision-Recall (AUPRC), and F-1 score as performance metrics. At inference, given new proteins and new compounds, DeepPurpose returns prediction scores representing predicted probabilities of binding between compounds and proteins.

\subsection{Modules for Other Downstream Prediction Tasks} 

DeepPurpose includes \texttt{repurposing} and \texttt{virtual\_screening} functions. Using only a few lines of codes that specify a list of compounds library to be screened upon and an optional set of training dataset, DeepPurpose trains five deep learning models, aggregates prediction results, and generates a descriptive ranked list in which compound candidates with the highest predicted binding scores are placed at the top. If the user does not specify a training dataset, DeepPurpose uses a pre-trained deep model for prediction. This list can then be examined to identify promising compound candidates for further experiments. Second, DeepPurpose also supports user-friendly programming frameworks for other modeling tasks, including drug and protein property prediction, drug-drug interaction prediction, and protein-protein interaction prediction (See Supplementary). Third, DeepPurpose provides an interface to many types of data, including public large binding affinity dataset~\citep{liu_bindingdb_2007}, bioassay data~\citep{kim2019pubchem}, and a drug repurposing library~\citep{corsello2017drug}.

%This loader can save users valuable time since the data are scattered and information to process the data is limited online. 

\subsection{Programming Framework and Implementation Details} 
The functionality of DeepPurpose is modularized into six key steps where a single line of code can invoke each step:  a) Load the dataset from a local file or load a DeepPurpose benchmark dataset. b) Specify the names of compound and protein encoders. c) Split the dataset into training, validation and testing sets using \texttt{data\_process} function, which implements a variety of data-split strategies. d) Create a configuration file and specify model parameters. If needed, DeepPurpose can automatically search for optimal values of hyper-parameters. e) Initialize a model using the configuration file. Alternatively, the user can load a pre-trained model or a previously saved model. f) Finally, train the model using \texttt{train} function and monitor the progress of training and performance metrics.  DeepPurpose is OS-agnostic and uses the Jupyter Notebook interface. It can be run in the cloud or locally. All datasets, models, documentation, installation instructions, and tutorials are provided at \href{https://github.com/kexinhuang12345/DeepPurpose}{https://github.com/kexinhuang12345/DeepPurpose}.

\section{Using DeepPurpose for DTI Prediction}

To demonstrate the use of DeepPurpose, we compare DeepPurpose with KronRLS~\citep{pahikkala_toward_2015}, a popular DTI method, and GraphDTA~\citep{nguyen_graphdta_2020} and DeepDTA~\citep{ozturk_deepdta_2018}, state-of-the-art DL methods. We find that many DeepPurpose models achieve comparable prediction performance on two benchmark datasets,  DAVIS~\citep{davis_comprehensive_2011} and KIBA~\citep{he_simboost_2017} (Figure~\ref{fig:my_label}D). A complete script to generate the results is provided in Supplementary Information.

\section{DeepPurpose with Interactive Web Interface}
In addition to rapid model prototyping, DeepPurpose also provides utility functions to load a pre-trained model and make predictions for a new drug and target inputs. This functionality allows domain scientists to examine predictions quickly, modify the inputs based on predictions, and iterate on the process until finding a drug or target with desired properties. We leverage Gradio~\citep{abid2019gradio} to create a web interface programmatically. We use a user-trained DeepPurpose model in the backend and create a custom web interface in fewer than ten code lines. This web interface takes the SMILES and amino acid sequence as the input and returns prediction scores with less than 1-second latency. We provide examples in the Supplementary.

\bibliographystyle{unsrtnat}
\bibliography{ref}

\appendix

\section{Further Details on DeepPurpose}

\paragraph{Overview of DeepPurpose Framework.} \mname uses an encoder-decoder framework for DTI prediction. The input of \mname is a compound SMILES string and protein amino acid sequence pair. The output of \mname is a score that measures the binding activity of the input compound protein pair. The power of deep learning models comes from its ability to create predictive feature vectors also called {\it embeddings}. In particular, \mname encodes both input compound and protein through various deep learning encoders to obtain their deep embeddings, then concatenates and feed them into a decoder, which is another deep neural network that aims to classify whether the input compound and target protein bind. 

\begin{figure}[t]
    \centering
    \includegraphics[width = 0.7 \textwidth]{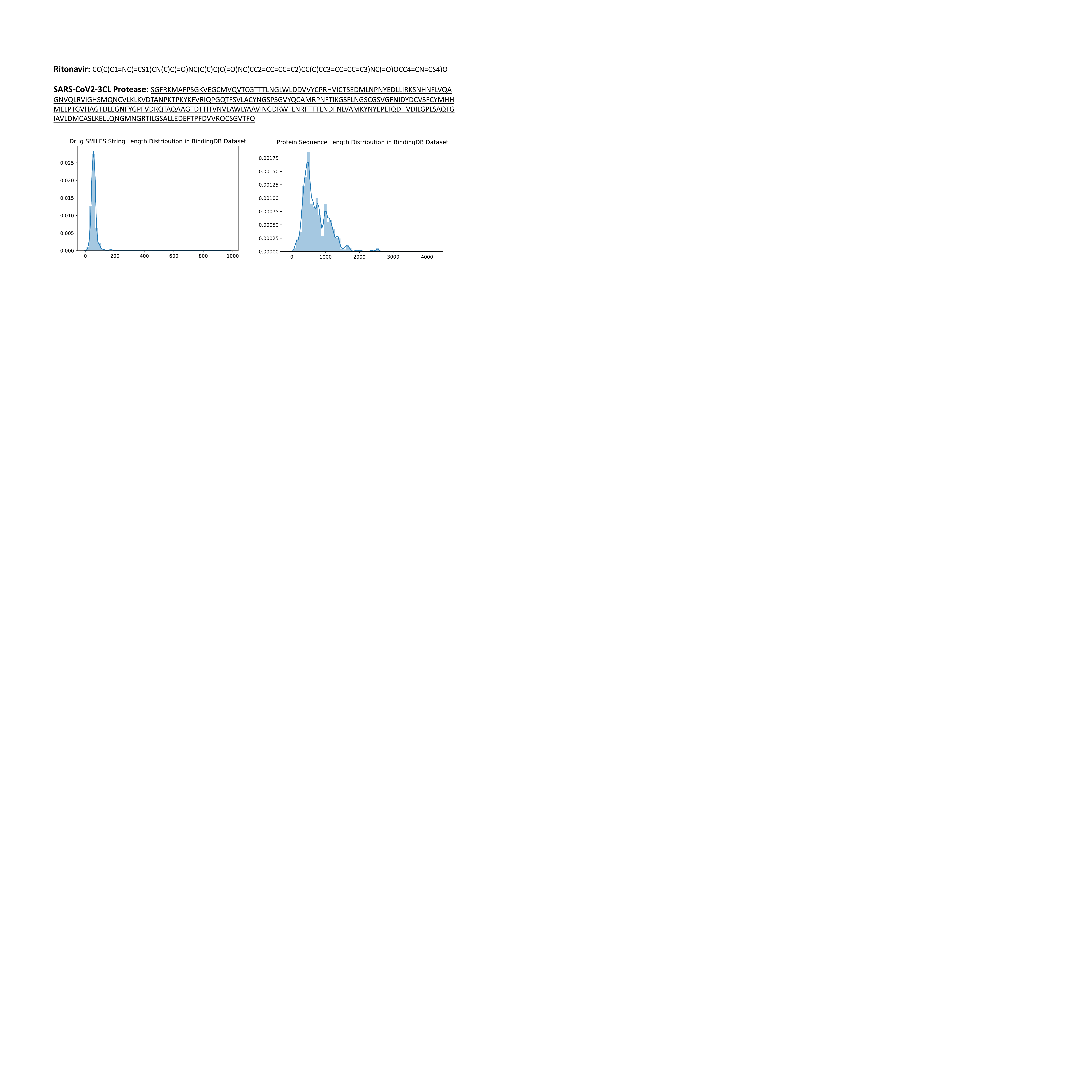}
    \caption{Examples of input representation. Compound is represented as SMILES string and protein is represented as amino acid sequence. The length distribution of the input are provided.}
    \label{fig:3}
\end{figure}

\paragraph{Encoders Implemented in DeepPurpose.} \mname provides 8 compound encoders and 7 protein encoders with numerous variants, ranging from classic chemical informatics fingerprints to various deep neural networks. \mname feed two latent vectors generated from compound and protein encoders into the decoder to produce the final prediction score. With such a pipeline design, switching encoders is very simple in \mname. By configuring a different encoder name \mname will automatically switch to the required encoder model and connect them with the decoder for  prediction. 

\paragraph{Compound Encoders.}
The input compound is represented by SMILES strings corresponding to molecule graphs~(Figure.~\ref{fig:3}).

\begin{enumerate}
    \item \textbf{Morgan Fingerprint}~\cite{rogers2010extended} is a 1024-length bits vector that encodes circular radius-2 substructures. A multi-layer perceptron is then applied on the binary fingerprint vector.
    \item \textbf{Pubchem}~\cite{kim2019pubchem} is a 881-length bits vector, where each bit corrresponds to a hand-crafted important substructures. A multi-layer perceptron is then applied on top of the vector.
    \item \textbf{Daylight}~\footnote{Daylight chemical information systems: https://www.daylight.com/} is a 2048-length vector that encodes path-based substructures. A multi-layer perceptron is then applied on top of the vector.
    \item \textbf{RDKit-2D}~\footnote{https:
//github.com/bp-kelley/descriptastorus} is a 200-length vector that describes global pharmacophore descriptor. It is normalized to make the range of the features in the same scale using cumulative density function fit given a sample of the molecules.
    \item \textbf{CNN}~\cite{krizhevsky2012imagenet} is a multi-layer 1D convolutional neural network. The SMILES characters are first encoded with an embedding layer and then fed into the CNN convolutions. A global max pooling layer is then attached and a latent vector describe the compound is generated.
    \item \textbf{CNN+RNN}~\cite{cho-etal-2014-learning,hochreiter1997long} attaches a bidirectional recurrent neural network (GRU or LSTM) on top of the 1D CNN output to leverage the more global temporal dimension of compound. The input is also the SMILES character embedding.
    \item \textbf{Transformer}~\cite{vaswani2017attention} uses a self-attention based transformer encoder that operates on the sub-structure partition fingerprint~\cite{huang2019caster}. 
    \item \textbf{MPNN}~\cite{gilmer2017neural} is a message-passing graph neural network that operate on the compound molecular graph. It transmits latent information among the atoms and edges, where the input features incorporate atom/edge level chemical descriptors and the connection message. After obtaining embedding vector for each atom and edge, a readout function (mean/sum) is used to obtain a (molecular) graph-level embedding vector. 
\end{enumerate}

\paragraph{Protein Encoders.} 
The input targets are proteins represented as sequences of 20 different kinds of amino acids~(Figure.~\ref{fig:3}). 
\begin{enumerate}
    \item \textbf{AAC}~\cite{reczko1994def} is a 8,420-length vector where each position correpsonds to an amino acid k-mers and k is up to 3. 
    \item \textbf{PseAAC}~\cite{chou2005using} includes the protein hydrophobicity and hydrophilicity patterns information in addition to the composition. 
    \item \textbf{Conjoint Triad}~\cite{shen2007predicting} uses the continuous three amino acids frequency distribution from a hand-crafted 7-letter alphabet.
    \item \textbf{Quasi Sequence}~\cite{chou2000prediction} takes account for the sequence order effect using a set of sequence-order-coupling numbers.
    \item \textbf{CNN}~\cite{krizhevsky2012imagenet} is a multi-layer 1D convolutional neural network. The target amino acid is decomposed to each individual character and is encoded with an embedding layer and then fed into the CNN convolutions. It follows a global max pooling layer.
    \item \textbf{CNN+RNN}~\cite{cho-etal-2014-learning,hochreiter1997long} attaches a bidirectional recurrent neural network (GRU or LSTM) on top of the 1D CNN output to leverage the sequence order information.
    \item \textbf{Transformer}~\cite{vaswani2017attention} uses a self-attention based transformer encoder that operates on the sub-structure partition fingerprint~\cite{huang2019caster} of proteins. Since transformer's computation time and memory is quadratic on the input size, it is computational infeasible to treat each amino acid symbol as a token. The partition fingerprint decomposes amino acid sequence into protein substructures of moderate sized such as motifs and then each of the partition is considered as a token and fed into the model.   
\end{enumerate}

\paragraph{Note on Feature-Architecture Combinations} 

During implementation, we notice that all of the architectures require specific input features, which means other input features are incompatible with the architecture. DeepPurpose currently supports five architectures: MLP, CNN, CNN+RNN, Transformer, and MPNN, and it supports five types of features: fingerprints, SMILES string, ESPF fingerprint, and molecular graph. Notice that these architectures are not suitable for alternative features. MLP can take in fingerprints vectors, but one-hot matrices of SMILES or graphs are incompatible. CNN/CNN+RNN expect a matrix where each row is a one-hot vector for an entity in the SMILES/Amino Acids, but cannot take a single fingerprint vector or graphs. Transformer is quadratic to the input length, which makes the long fingerprints and SMILES/Amino acids computationally expensive. MPNN operates on graphs, thus, it is incompatible with fingerprints, SMILES, ESPF. In addition, the current included features \& encoders combinations are not randomly assembled but are previously included in the literature and have shown strong performance for molecular modeling with compounds and targets.

\paragraph{Programming Framework}
The functionality of DeepPurpose is modularized into six key steps where a single line of code can invoke each step: a) Load the dataset from a local file or load a DeepPurpose benchmark dataset. b) Specify the names of compound and protein encoders. c) Split the dataset into training, validation and testing sets using \texttt{data\_process} function, which implements a variety of data-split strategies. d) Create a configuration file and specify model parameters. If needed, DeepPurpose can automatically search for optimal values of hyper-parameters. e) Initialize a model using the configuration file. Alternatively, the user can load a pre-trained model or a previously saved model. f) Finally, train the model using \texttt{train} function and monitor the progress of training and performance metrics.

\paragraph{Downstream Prediction, Objective Function, and Inference.} After \mname obtains latent compound and protein embedding, both are fed into a multi-layer perceptron decoder. There are two classes of tasks/datasets in drug target interaction prediction. One's label is binding score such as Kd, IC50, and they are continuous values while the other one's label is binary, whether or not they can bind. \mname is able to automatically detect whether the task is regression for continuous label or classification for binary label by counting the number of unique labels in the data. For binding affinity score prediction, it uses mean squared error (MSE) loss (Eq.~\ref{eq:MSE}). For binary interaction prediction, it uses binary cross entropy (BCE) loss (Eq.~\ref{eq:BCE}).
\vspace{-5mm}

\begin{equation}\label{eq:MSE}
    \mathcal{L}_{MSE} = \frac{1}{n} \sum_{i = 1}^n \left( \mathrm{y}_\mathrm{i} - \hat{\mathrm{y}}_\mathrm{i} \right)^2
\end{equation}

\vspace{-7mm}
\begin{equation}\label{eq:BCE}
    \mathcal{L}_{BCE} = \frac{1}{n} \sum_{i = 1}^n  \mathrm{y}_\mathrm{i} \mathrm{log}\left(\hat{\mathrm{y}}_\mathrm{i}\right) + \left(1 - \mathrm{y}_\mathrm{i}\right) \mathrm{log}\left(1 - \hat{\mathrm{y}}_\mathrm{i}\right),
\end{equation}
where $\mathrm{y}_\mathrm{i}$ is the true label and $\hat{\mathrm{y}}_\mathrm{i}$ is the predicted label for $\mathrm{i}$-th compound-protein pair. For evaluation metrics, we use MSE, Concordance Index, and Pearson Correlation for continuous regression and Receiver Operating Characteristics-Area Under the Curve (ROC-AUC), Precision Recall-Area Under the Curve (PR-AUC) and F1 score at threshold 0.5 for binary classification. During inference, given new proteins or new compounds, the model prediction is used as the predicted binding score/interaction probability.

\section{DeepPurpose for Other Tasks Related to Compound and Protein} 

\begin{figure}
    \centering
    \includegraphics[width=\textwidth]{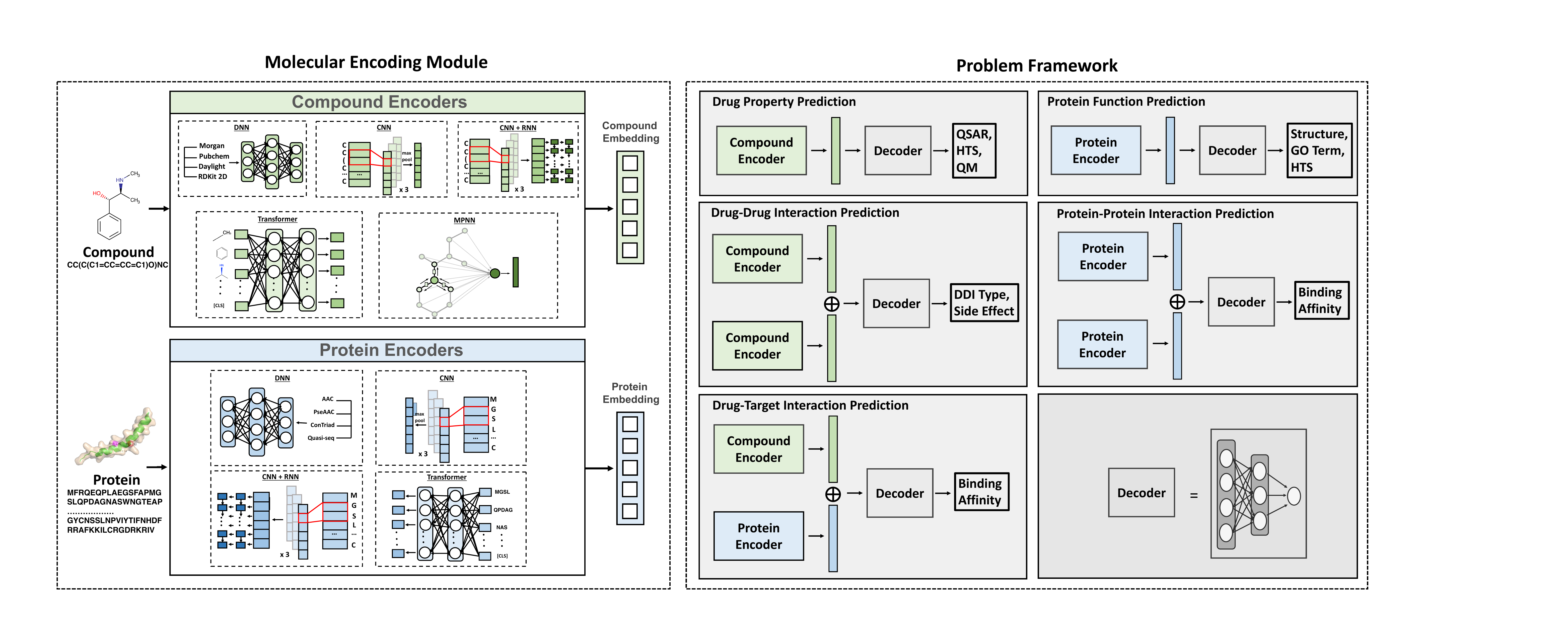}
    \caption{DeepPurpose supports DTI, DDI, PPI, Drug Property and Protein Function Prediction tasks.}
    \label{fig:frameworks}
\end{figure}

In addition to DTI prediction, DeepPurpose also supports user-friendly programming frameworks for other molecular modeling tasks, namely, drug/protein property prediction, drug-drug interaction prediction, protein-protein interaction prediction tasks. We provide a framework illustration in Figure.~\ref{fig:frameworks}. 

\clearpage

\appendix

\end{document}

%% file: math_commands.tex
\usepackage{amsmath,amsfonts,bm}

\def\eqref#1{equation~\ref{#1}}

\DeclareMathAlphabet{\mathsfit}{\encodingdefault}{\sfdefault}{m}{sl}
\SetMathAlphabet{\mathsfit}{bold}{\encodingdefault}{\sfdefault}{bx}{n}